%% file: ms.tex
\documentclass[12pt]{report}

\usepackage{aaai-bib}
\usepackage{afterpage}
\usepackage{algorithm}
\usepackage{algpseudocode}
\usepackage{amsfonts}
\usepackage{amsmath}
\usepackage{booktabs}
\usepackage{courier}
\usepackage{dsfont}
\usepackage{epsfig}
\usepackage[hang,flushmargin]{footmisc} 
\usepackage[margin=1in]{geometry}
\usepackage{graphicx}
\usepackage{helvet}
\usepackage{multirow}
\usepackage{natbib}
\usepackage[all]{nowidow}
\usepackage{scrextend}
\usepackage[doublespacing]{setspace}		
\usepackage{subcaption}
\usepackage{times}
\usepackage[table, dvipsnames]{xcolor}

\graphicspath{{./img/}}

\definecolor{table-gray}{gray}{0.96}

\newcommand{\E}[1]{{\mathbb E}\left[ #1 \right]}
\newcommand{\norm}[1]{\left\lVert#1\right\rVert}
\newcommand{\tworow}[2][c]{
	\begin{tabular}[#1]{@{}c@{}}#2\end{tabular}
}
\newcommand{\inserttitle}{ELASTIC GOSSIP: DISTRIBUTING NEURAL NETWORK TRAINING USING GOSSIP-LIKE PROTOCOLS}
\newcommand{\insertauthor}{Siddharth Pramod}
\newcommand{\insertdegree}{MS, Computer Science}
\newcommand{\insertyear}{2018}
\newcommand{\insertadvisorname}{Tim Oates}
\newcommand{\insertadvisortitle}{Professor}

\input{psfig.tex}
\def\fchaptermark#1{}

\begin{document}
\title{{\bf \inserttitle}}
\author{\insertauthor}
\tolerance=1000
\newpage
\include{abstract}

\newpage
\include{titlepage}
\newpage
\include{copyright}

\pagenumbering{roman}

\newpage
\setcounter{page}{2}
\cleardoublepage
\newpage
\include{dedication}
\cleardoublepage
\include{acknowledgment}
\cleardoublepage
\tableofcontents
\cleardoublepage
\listoftables
\cleardoublepage
\listoffigures
\cleardoublepage

\pagenumbering{arabic}
\pagestyle{myheadings}
\markright{}

\include{./chapters/1_introduction}
\include{./chapters/2_related_work}
\include{./chapters/3_elastic_gossip}
\include{./chapters/4_experiments}
\include{./chapters/5_conclusion}

\cleardoublepage
\appendix
\include{./chapters/appendix}

\cleardoublepage

\thispagestyle{plain}
\bibliographystyle{aaai}
\bibliography{bib}

\end{document}

%% file: abstract.tex
\newpage
\pagestyle{empty}

\begin{center}
\vspace{0.1in}
\large{\bf ABSTRACT} \par  
\bigskip \bigskip
\end{center}

\begin{flushleft}
{\bf Title of Thesis:} \inserttitle\\
\insertauthor, \insertdegree, \insertyear \\
\begin{singlespace}
{\bf Thesis directed by:}{\hspace{2.5mm}} \parbox[t]{3in}{\insertadvisorname, \insertadvisortitle \\
Department of Computer Science and \\ Electrical Engineering}
\end{singlespace}
\end{flushleft}

Distributing Neural Network training is of particular interest for several reasons including scaling using computing clusters, training at data sources such as IOT devices and edge servers, utilizing underutilized resources across heterogeneous environments, and so on. Most contemporary approaches primarily address scaling using computing clusters and require high network bandwidth and frequent communication. This thesis presents an overview of standard approaches to distribute training and proposes a novel technique involving pairwise-communication using Gossip-like protocols, called Elastic Gossip. This approach builds upon an existing technique known as Elastic Averaging SGD (EASGD), and is similar to another technique called Gossiping SGD which also uses Gossip-like protocols. Elastic Gossip is empirically evaluated against Gossiping SGD using the MNIST digit recognition and CIFAR-10 classification tasks, using commonly used Neural Network architectures spanning Multi-Layer Perceptrons (MLPs) and Convolutional Neural Networks (CNNs). It is found that Elastic Gossip, Gossiping SGD, and All-reduce SGD perform quite comparably, even though the latter entails a substantially higher communication cost. While Elastic Gossip performs better than Gossiping SGD in these experiments, it is possible that a more thorough search over hyper-parameter space, specific to a given application, may yield configurations of Gossiping SGD that work better than Elastic Gossip.
\par\vfil

%% file: titlepage.tex
\begin{titlepage}
\mbox{}\vspace{1in}
\begin{center}

    {\Large \bf \inserttitle \par}
    
\vspace{1.7in}

    {\large by} \\
    {\large \insertauthor}
    
\vspace{1.7in}

  \begin{singlespace}
    Thesis submitted to the Faculty of the Graduate School \\
    of the University of Maryland in partial fulfillment \\
    of the requirements for the degree of \\
    Master of Science, Computer Science \\
    2018
	\end{singlespace}
\end{center}
\end{titlepage}

%% file: copyright.tex
\begin{titlepage}
\mbox{}\vspace{7.5in}
\begin{center}
\copyright~Copyright \insertauthor \insertyear
\end{center}
\end{titlepage}

%% file: dedication.tex
\newpage
\thispagestyle{plain}
\vfil\null
\begin{center}

\mbox{}\vspace{3in}

\emph{Dedicated to my mother, Vinita, and my wife, Priyanka, for their endless support in all of my endeavors.}

\end{center} 
\normalsize
\vfil\null

%% file: acknowledgment.tex
\section*{ACKNOWLEDGMENTS}
\pagestyle{plain}

I sincerely thank my thesis advisor, Dr. Tim Oates, for his invaluable guidance in conducting research and his continued support throughout my time as a Masters student. I am very grateful to Dr. Kostas Kalpakis and Dr. Mohamed Younis for being part of my thesis examination committee and for all the great feedback provided. I thank my lab-mate, Karan Budhraja, for the very insightful discussions we have had contributing to many of the ideas in this thesis. I thank my former colleague, Neeraj Kashyap, for assisting me with setting up experiments on Google Cloud Platform. I thank my wife, Priyanka, who has stood by me at every step during this course.

\clearpage

%% file: chapters/1_introduction.tex
\chapter{INTRODUCTION}
\thispagestyle{plain}

\label{ch:intro}

There has been considerable interest in distributing neural network training for several reasons. Primarily, deep neural networks are inherently well suited to making use of large volumes of training data, while simultaneously being computationally expensive, relative to other machine learning techniques. Thus, one approach to scaling has been through distributed computation. Additionally, data sources for a single application are often many. Due to the speed and volume at which data is collected, one may wish to collocate training with data collection (as with data processing), so as to avoid the cost of consolidating raw data when technically feasible. Further, distributed training may be a requirement where the training data itself may not be transferable due to privacy concerns, such as those involving patients' health data in federated systems. \\

The most commonly used techniques to distribute neural network training employ distributed variants of Stochastic Gradient Descent (SGD). They may broadly be categorized as model-parallel or data-parallel, in reference to whether the neural network model or training data, respectively, are partitioned across nodes \citep{dean2012downpour}. They may also be categorized as synchronous or asynchronous, owing to communication constraints between nodes of the distributed system, as required by the respective technique. Details on these forms of categorization are expounded upon in Chapter \ref{ch:related-work}. \\

Efforts towards distributing neural network training have, thus far, focused on scaling existing applications to reduce training time. Accordingly, such approaches are evaluated based on how they scale while approaching or surpassing state-of-the-art results on standard benchmarks. On the other hand, while there is interest in understanding training behavior in inherently distributed systems such as wireless sensor networks and IOT devices, there appears to be very little information about this in the public domain. The goal of this thesis is to address this gap. To this end, a \textit{decentralized} approach to training neural networks, called \textit{Elastic Gossip}, is proposed, details of which are provided in Chapter \ref{ch:elastic-gossip}. This approach extends \textit{Elastic Averaging SGD} (EASGD) \citep{zhang2015easgd}, and is similar to \textit{Gossiping SGD} \citep{jin2016gossip} and \textit{GoSGD} \citep{blot2016gossip}, utilizing gossip-like protocols for \textit{synchronous} communication of model parameters between nodes in a data-parallel setting. While Elastic Gossip is not the only gossip-like approach that can be formulated as an extension of EASGD, it is the first one (and only one as of this writing) that maintains \textit{elastic symmetry} in updates. \\

Evaluations of Elastic Gossip against Synchronous All-reduce SGD, and Gossiping SGD specifically in the synchronous setting are discussed in Chapter \ref{ch:experiments}. The latter evaluation runs contrary to the original work on Gossiping SGD that used an asynchronous setting, as the purpose then was to study scaling. However, experimental results in asynchronous settings are subject to extraneous factors such as computing hardware, environment, other concurrently running processes, communication networks, etc., and so this thesis studies only synchronous settings in the interest of reproducibility. The evaluations are performed using the MNIST and CIFAR-10 benchmarks. We do not evaluate Elastic Gossip against Elastic Averaging SGD itself as the latter requires a central process that communicates with all of the other workers, and therefore may not be suitable for decentralized training. \\

This thesis also lays some groundwork for future studies in the space of decentralized distributed training using protocols that are aware of underlying network topologies and associated costs, the effects of biases and skews in data distribution across nodes, and studying the effects of asynchrony that is controlled in a simulated environment.

%% file: chapters/2_related_work.tex
\chapter{Background and Related Work}
\thispagestyle{plain}
\label{ch:related-work}

The work presented in this thesis most closely relates to Elastic Averaging SGD \citep{zhang2015easgd}, Gossiping SGD \citep{jin2016gossip} and GoSGD \citep{blot2016gossip}. These are discussed in detail in Sections \ref{sec:easgd} and \ref{sec:gossiping-sgd}. A brief overview of the distributed deep learning landscape is provided in Section \ref{sec:background}. \\

The following discussion will use the notion of a process as an abstraction. Every process is expected to be a stand-alone entity that can nevertheless communicate with other processes taking part in neural network training. These processes can be hosted either on a single machine, a distributed system, or a combination. Processes take on one of two roles - (1) worker processes that train a given model or partition of the model, (2) central processes that co-ordinate training among the worker processes. \\

\section{Background}
\label{sec:background}

Distributed deep learning techniques are often classified as model-parallel or data-parallel \citep{dean2012downpour, yadan2013multigpu}, a classification that applies to multi-gpu, multi-machine, or multi-process architectures in general. Model-parallelism entails partitioning a given neural network model across multiple processes, with each process updating only the associated parameters. \textit{AlexNet} \citep{krizhevsky2012alexnet} is a popular multi-GPU architecture that employs model-parallelism. On the other hand, data-parallelism entails partitioning training data across multiple processes, with each process receiving a replica of the entire model to train. Frameworks such as \textit{DistBelief} \citep{dean2012downpour} allow incorporation of hybrid techniques that make use of model- and data-parallelism. While model-parallelism generally incurs higher communication costs, it is often used when hosting an entire model on a single process is infeasible, for example, due to hardware constraints. Distributed deep learning techniques may also be classified as synchronous or asynchronous, depending on whether the associated algorithm enforces synchronization between the processes. A third form of categorization is based on whether the given technique requires the use of a central co-ordinating process, typically termed the parameter server, as its task is primarily to maintain a central repository of parameters with which all other processes are periodically synchronized. Note that some frameworks such as \textit{DistBelief} make use of a distributed parameter server in order to scale, however, its function remains the same. \\

Elastic gossip, as presented in this thesis, is a data-parallel technique that does not make use of parameter servers. While the algorithms and experiments presented here only use the synchronous setting, Elastic Gossip can be trivially extended to the asynchronous setting as will be explained in Chapter \ref{ch:elastic-gossip}. \\

\subsection{Synchronous All-reduce SGD}
Synchronous All-reduce SGD, hereafter referred to as \textit{All-reduce SGD}, is an extension of Stochastic Gradient Descent purposed for distributed training using a data-parallel setting. At each training step, gradients are first computed using backpropagation at each process, sampling data from the partition it is assigned. The gradients are then aggregated across all processes, with each process receiving a copy of this aggregate. The aggregate itself is either a sum or an average, depending on whether gradients are summed or averaged across training data instances respectively. The parameters at each process are then updated using this aggregated gradient. The update rules are more formally given in Algorithm \ref{alg:sync-all-reduce}. \\

The aggregation step in Line \ref{alg-line:sync-all-reduce:all-reduce} of Algorithm \ref{alg:sync-all-reduce} constitutes the \textit{all-reduce} operation - a term derived from the \textit{reduce} operation used in functional programming, and the fact that \textit{all} workers have access to the result of the operation. There are several system architectures that have been proposed for All-reduce SGD. These initially utilized a central process to communicate with all worker processes individually, and was responsible for the all-reduce operation. This was made more efficient using reduction-tree based algorithms \citep{iandola2016firecaffe}. Most recently, ring-based all-reduce has been proposed as an alternative that does not require a central process, and is also more communication-efficient in that data transferred to and from each process is independent of the number of processes in the system \citep{amodei2015deepspeech2,patarasuk2009allreduce,thakur2005allreduce}. \\

\begin{algorithm}
	\caption{Synchronous All-reduce SGD}
	\label{alg:sync-all-reduce}
	\begin{algorithmic}
		\State \textbf{Inputs:}
			\State \quad $\mathcal{X}$ \Comment{\parbox[t]{.8\linewidth}{training data instances}}
			\State \quad $\theta$ \Comment{\parbox[t]{.8\linewidth}{initial parameters of the model}}
			\State \quad $f(\theta; \mathcal{X})$ \Comment{\parbox[t]{.8\linewidth}{loss function}}
			\State \quad $W$ \Comment{\parbox[t]{.8\linewidth}{set of worker processes}}
			\State \quad $\eta$ \Comment{\parbox[t]{.8\linewidth}{learning rate}}
		\State \textbf{Initialize:}
			\State \quad $\theta^i \leftarrow \theta$, for all $i \in W$
			\State \quad $t^i \leftarrow 0$, for all $i \in W$
	\end{algorithmic}
	\hrulefill
	\begin{algorithmic}[1]
		\Repeat \: for all $i \in W$ in parallel
			\State $g^i \leftarrow \nabla f(\theta^i; x \small{\sim} \mathcal{X}^i)$
				\Comment{\parbox[t]{.53\linewidth}{compute gradients}}
			\State Wait until $t^i = t^j$, for all $j \in W, i \ne j$	
				\Comment{\parbox[t]{.33\linewidth}{synchronize}}
			\State $g^i \leftarrow \frac{1}{\left|W\right|}\sum_{k \in W}(g^k)$
				\Comment{\parbox[t]{.5235\linewidth}{all-reduce on gradients}}
				\label{alg-line:sync-all-reduce:all-reduce}
			\State $\theta^i \leftarrow \theta^i - \eta g^i$
				\Comment{\parbox[t]{.53\linewidth}{update parameters}}
			\State $t^i \leftarrow t^i + 1$
				\Comment{\parbox[t]{.53\linewidth}{update clock}}
		\Until forever
	\end{algorithmic}
\end{algorithm}

Note that barring any distinctions in sampling due to data distribution, All-reduce SGD is mathematically equivalent to Stochastic Gradient Descent with mini-batches, where, if $W$ is the set of workers in the system, then the effective batch size is $|W|$ times the batch-size used at each worker. \\

\subsection{Motivating asynchrony}
At the synchronization point in All-reduce SGD, each worker has to wait for every other worker in the distributed system to finish computing gradients before the all-reduce step. Therefore, training time is constrained by the slowest processes (``stragglers'') in the system. This criticism is true for synchronous methods in general. While there has been at least one attempt to alleviate this concern using redundancy \citep{chen2016backup}, several researchers have instead sought out asynchronous alternatives for distributed neural network training. \\

Early asynchronous techniques utilized a central parameter store and multiple worker processes, where parameters in the central store were updated by the workers in a lock-free manner. The first among these was a single-machine approach called \textit{Hogwild!} \citep{recht2011hogwild}, followed by a distributed extension called \textit{Downpour} \citep{dean2012downpour}. Hogwild! maintains the parameter-store in shared memory, while Downpour uses a parameter server. In both cases, each worker retrieves a copy of the parameters from the parameter store, computes updates, and writes them back to the parameter-store. The implication of being lock-free is that some processes can write updates based on gradients that were computed using a parameter state that might since have become stale. This is shown to be viable when gradients are generally sparse, as is often true with neural network training, such that every process concurrently writing to the store updates a near-exclusive set of parameters \citep{recht2011hogwild}. Additionally, some researchers have attempted to account for staleness using staleness-aware learning rates \citep{zhang2015staleness}. To reduce communication overhead, there has been at least one attempt to quantize updates \citep{strom2015quantized}, building on the idea that gradients are sparse. \\

While there is good reason to develop asynchronous approaches to training neural networks, it is difficult to conduct reproducible studies due to the effect extraneous factors have on asynchrony. Thus, this thesis focuses on synchronous variants in the experiments, even though it builds upon previously proposed asynchronous techniques. Using the results of synchronous formulations as a basis, the asynchronous variants can also be studied in a controlled manner through simulation. \\

\section{Elastic Averaging SGD}
\label{sec:easgd}

\citet{zhang2015easgd} propose a novel asynchronous data-parallel SGD technique called \textit{Elastic Averaging SGD} (EASGD), aimed at reducing communication overhead, while simultaneously addressing the issue of gradient staleness described earlier. The architecture consists of a set of worker processes $W$, and a single central process that communicates with all of the workers, in a data-parallel setting. The central process is similar to a parameter server, except that it plays a role in the mathematical formulation for training using EASGD, wherein it maintains the \textit{consensus} - an aggregate of parameters from each of the workers. \\

The optimization problem is formulated as jointly minimizing training loss at each worker process, while simultaneously penalizing workers for deviating from the consensus in parameter-space. This joint objective function is shown in Equation \ref{eq:easgd-objective}, and is based on the \textit{global consensus problem} discussed by \citet{boyd2011admm}.

\begin{equation}
	\label{eq:easgd-objective}
	\min_{\tilde{\theta}, \theta_i \forall i \in W} \sum_{i \in W} \left[ \: \E{f(\theta^i; \mathcal{X}^i)} + \frac{\rho}{2} \norm{\theta^i - \tilde{\theta}}_2^2 \: \right]
\end{equation}

\noindent where $W$ is the set of worker processes, $\theta^i$ are the workers' replicas of the model parameters, referred to as \textit{local variables}, $\tilde{\theta}$ is the consensus, referred to as the \textit{center variable}, $\mathcal{X}^i$ are the partitions of training data, and $f$ is the learning objective. The first term under the summation is the training loss at each worker process, and the second term is a quadratic penalty on the disagreement between $\theta^i$ and $\tilde{\theta}$. $\tilde{\theta}$ can be thought of as representing all $\theta^j, \forall j \in W$. The weight on the quadratic penalty term, $\rho$, is a hyper-parameter that determines the degree to which worker processes are allowed to explore the parameter space by deviating from the consensus, and simultaneously determines how much the center variable is influenced by the local variables at any given time. At one extreme, $\rho = 0$ would result in the workers learning solely from the partition of training data that it is assigned, unconstrained by the consensus (and thus any other worker), while the central process does not perform any update at all. As $\rho$ increases, the worker processes are allowed to explore less using their partitions, but are ``aided'' by what other workers have learned thus far, as represented by the consensus. \\

Taking derivatives of the objective function in Equation \ref{eq:easgd-objective} w.r.t $\theta^i$ and $\tilde{\theta}$, the corresponding update rules based on Stochastic Gradient Descent are shown in Equations \ref{eq:easgd-update1} and \ref{eq:easgd-update2} respectively \citep{zhang2015easgd}.

\begin{equation}
	\label{eq:easgd-update1}
	\theta^i_{t+1} = \theta^i_t - \eta \: \nabla f \! \left( \theta^i_t \right) - \alpha \left( \theta^i_t - \tilde{\theta}_t \right) \\
\end{equation}
\begin{equation}
	\label{eq:easgd-update2}
	\tilde{\theta}_{t+1} = \left(1 - \beta \right) \tilde{\theta}_t + \beta \left( \frac{1}{\left|W\right|} \sum_{i \in W} \theta^i_t \right)
\end{equation}

\noindent where $\eta$ is the learning rate, $\alpha=\eta\rho$ and $\beta=\eta^\prime\rho$, such that $\eta^\prime$ can be a learning rate different from $\eta$. Explicitly choosing $\beta=\alpha |W|$ results in an \textit{elastic symmetry} in updates: $|\alpha ( \theta^i_t - \tilde{\theta}_t )|$, which is shown to be crucial for the algorithm's stability \citep{zhang2015easgd}. Accordingly, the center variable update in Equation \ref{eq:easgd-update2} is rewritten as shown in Equation \ref{eq:easgd-update3}. Through the rest of this thesis, $\alpha$ is referred to as the \textit{moving rate} and is a hyper-parameter used in EASGD as well as in Elastic Gossip.

\begin{equation}
	\label{eq:easgd-update3}
	\tilde{\theta}_{t+1} = \tilde{\theta}_t + \alpha \sum_{i \in W} \left( \theta^i_t - \tilde{\theta}_t \right)
\end{equation}

Algorithm \ref{alg:easgd-sync} presents update rules for Synchronous EASGD, which is very similar to the asynchronous version discussed in the original work, except that the clocks $t^i$ are not synchronized in the latter. Note that there are two forms of updates to the parameters: (1) those corresponding to gradients (2) those involving communication. This is a common pattern among techniques discussed in this thesis, and will be referred to as the gradient-related and communication-related components respectively. Also note that communication is restricted to every $\tau$ updates instead of every single update as is shown in the equations, where $\tau$ is termed the communication period. Besides reducing communication overhead, $\tau$ can also be used to control the explore-exploit trade-off along with $\alpha$. \\

\begin{algorithm}
	\caption{Synchronous EASGD}
	\label{alg:easgd-sync}
	\begin{algorithmic}
		\State \textbf{inputs:}
			\State \quad $\mathcal{X}$ \Comment{\parbox[t]{.8\linewidth}{training data instances}}
			\State \quad $\theta$ \Comment{\parbox[t]{.8\linewidth}{initial parameters of the model}}
			\State \quad $f(\theta; \mathcal{X})$ \Comment{\parbox[t]{.8\linewidth}{loss function}}
			\State \quad $W$ \Comment{\parbox[t]{.8\linewidth}{set of worker processes}}
			\State \quad $\eta$ \Comment{\parbox[t]{.8\linewidth}{learning rate}}
			\State \quad $\alpha$ \Comment{\parbox[t]{.8\linewidth}{moving rate}}
			\State \quad $\tau$ \Comment{\parbox[t]{.8\linewidth}{communication period}}
		\State \textbf{initialize}
			\State \quad $\tilde{\theta} \leftarrow \theta$
			\State \quad $\theta^i \leftarrow \theta$, for all $i \in W$
			\State \quad $t^i \leftarrow 0$, for all $i \in W$
	\end{algorithmic}
	\hrulefill
	\begin{algorithmic}[1]
		\Repeat \: for all $i \in W$ in parallel
			\State $g^i \leftarrow \nabla f(\theta^i; x \small{\sim} \mathcal{X}^i)$
				\Comment{\parbox[t]{.53\linewidth}{compute gradients using the current state of parameters}}
			\If{$\tau$ divides $t^i$}
				\State Wait until $t^i = t^j$, for all $j \in W, i \ne j$	
					\Comment{\parbox[t]{.33\linewidth}{synchronize}}
				\State $z^i \leftarrow \alpha (\theta^i - \tilde{\theta})$
					\Comment{\parbox[t]{.53\linewidth}{compute the elastic update using the current state of parameters}}
				\State $\theta^i \leftarrow \theta^i - z^i$
					\Comment{\parbox[t]{.53\linewidth}{update local variable}}
				\State $\tilde{\theta} \leftarrow \tilde{\theta} + z^i$	\label{alg-line:center-update}
					\Comment{\parbox[t]{.53\linewidth}{update center variable}}
			\EndIf
			\State $\theta^i \leftarrow \theta^i - \eta g^i$
				\Comment{\parbox[t]{.53\linewidth}{update parameters}}
			\State $t^i \leftarrow t^i + 1$
				\Comment{\parbox[t]{.53\linewidth}{update clock}}
		\Until forever
	\end{algorithmic}
\end{algorithm}

Elastic Gossip extends Elastic Averaging SGD, where consensus is estimated using gossip-based protocols, eliminating the need for a center variable, thereby removing the communication bottleneck and using one less process. Additionally, Elastic Gossip can be deployed where decentralized training is a requirement. Elastic Gossip also maintains elastic symmetry in updates, motivated by its use and role in EASGD is not the only gossip-like approach that can be formulated as an extension of EASGD. \\

Note that Synchronous EASGD can be implemented by maintaining a copy of the center variable at each worker and updating it using an implementation of all-reduce that does not make use of a central process. This alternative eliminates the communication bottleneck associated with the central process, at twice the cost of storing parameters (both the local and center variables) at each worker. \\

\section{Gossiping SGD and GoSGD}
\label{sec:gossiping-sgd}

\citet{jin2016gossip} introduce Gossiping SGD, which uses gossip to estimate the communication-related component \citep{kempe2003gossip}, similar to EASGD. However, unlike EASGD, Gossiping SGD does not require a central process. The formulation for Gossiping SGD is shown to be related to EASGD, where the consensus represented by the center variable in the latter is replaced by an average of local variables in the former. This average is then estimated throughout the training process using gossip-like protocols. This is a very interesting approach that does not make use of a parameter server which is otherwise common among asynchronous approaches. Thus, Gossiping SGD constitutes decentralized training. \\ 

The update rules for the \textit{pull} variant of Gossiping SGD is presented in Algorithm \ref{alg:pull-gossiping-sgd}. A similar \textit{push} variant is presented in Algorithm \ref{alg:push-gossiping-sgd} in the Appendix. Note that these have been modified from the original to enforce synchrony. Additionally, while the communication-related and gradient-related updates are performed sequentially in the algorithms proposed originally, those presented here have been modified to compute them simultaneously, primarily to be consistent with the corresponding steps in EASGD \citep{zhang2015easgd}. Note that besides the communication-related component, these algorithms are identical to each other and to EASGD. \\

\begin{algorithm}
	\caption{Synchronous Pull-Gossiping SGD}
	\label{alg:pull-gossiping-sgd}
	\begin{algorithmic}
		\State \textbf{inputs:}
			\State \quad $\mathcal{X}$ \Comment{\parbox[t]{.8\linewidth}{training data instances}}
			\State \quad $\theta$ \Comment{\parbox[t]{.8\linewidth}{initial parameters of the model}}
			\State \quad $f(\theta; \mathcal{X})$ \Comment{\parbox[t]{.8\linewidth}{loss function}}
			\State \quad $W$ \Comment{\parbox[t]{.8\linewidth}{set of worker processes}}
			\State \quad $\eta$ \Comment{\parbox[t]{.8\linewidth}{learning rate}}
			\State \quad $\tau$ \Comment{\parbox[t]{.8\linewidth}{communication period}}
		\State \textbf{initialize}
			\State \quad $\tilde{\theta} \leftarrow \theta$
			\State \quad $\theta^i \leftarrow \theta$, for all $i \in W$
			\State \quad $t^i \leftarrow 0$, for all $i \in W$
	\end{algorithmic}
	\hrulefill
	\begin{algorithmic}[1]
		\Repeat \: for all $i \in W$ in parallel
			\State $g^i \leftarrow \nabla f(\theta^i; x \small{\sim} \mathcal{X}^i)$
				\Comment{\parbox[t]{.53\linewidth}{compute gradients using the current state of parameters}}
			\If{$\tau$ divides $t^i$}
				\State Wait until $t^i = t^j$, for all $j \in W, i \ne j$	
					\Comment{\parbox[t]{.33\linewidth}{synchronize}}
				\State $k \sim W \setminus \{i\}$
					\Comment{\parbox[t]{.53\linewidth}{select a peer at random}}
				\State $\theta^i \leftarrow \frac{1}{2}\left(\theta^i + \theta^k\right)$
					\Comment{\parbox[t]{.53\linewidth}{update parameters}}
			\EndIf
			\State $\theta^i \leftarrow \theta^i - \eta g^i$
				\Comment{\parbox[t]{.53\linewidth}{update parameters}}
			\State $t^i \leftarrow t^i + 1$
				\Comment{\parbox[t]{.53\linewidth}{update clock}}
		\Until forever
	\end{algorithmic}
\end{algorithm}

\citet{blot2016gossip} propose an alternate formulation for gossip-based training called GoSGD, which differs in a few ways from Gossiping SGD in the communication-related component. First, the updates are formulated based on the push-sum protocol proposed by \citet{kempe2003gossip}, such that the processes would converge to computing the average of parameters in the absence of gradient-related updates. Second, a communication probability $p$ is used instead of the communication period $\tau$ that's used in EASGD and Gossiping SGD, such that each process decides to communicate in a given iteration with probability $p$ of sampling $True$ from a Bernoulli Distribution. The communication period is then $1/p$ in expectation. This introduces stochasticity in the communication schedule. This is especially significant in the synchronous case, as communication schedules are then spread out across updates instead of workers communicating concurrently, potentially taxing communication networks. \\

Elastic Gossip is similar to Gossiping SGD, but differs in that it incorporates elastic symmetry in updates, motivated by the idea that it was shown to be crucial to stability of EASGD by \citet{zhang2015easgd}. Additionally, while in Gossiping SGD, the communication-related component is constrained to computing parameter averages, the same in Elastic Gossip incorporates a moving rate $\alpha$. Similar to its role in EASGD, this hyper-parameter represents the penalty on deviation from the consensus, and can be used to tweak the degree of exploration in parameter space. Elastic Gossip differs from GoSGD in terms of formulation, as the latter is not an extension of EASGD, although a more general formulation is discussed in Chapter \ref{ch:elastic-gossip}, from which each of EASGD, Gossiping SGD, GoSGD and Elastic Gossip may be derived. \\

\section{Existing empirical evaluations of approaches discussed}

\citet{zhang2015easgd} show empirically that asynchronous EASGD outperforms Downpour \citep{dean2012downpour}, both in terms of final test error and in time to converge. These experiments were conducted on the CIFAR-10 and ImageNet benchmarks using the architecture used by \citet{sermanet2014overfeat}, across multiple values of communication period, $\tau$, with the number of worker processes $|W| \in \{4, 8, 16\}$. Additionally, it is noted that using a larger number of worker processes correlates with convergence to a lower test error for EASGD, potentially related to the exploration-exploitation trade off in the parameter space. It is to be noted, however, that results may vary based on extraneous environmental factors as the experiments were conducted in an asynchronous setting. \\

\citet{jin2016gossip} evaluate Gossiping SGD against EASGD and All-reduce SGD, with experiments conducted on the Imagenet task using the Resnet-18 architecture \citep{he2016resnet}, over varying cluster sizes. They find that with 8 worker processes and $\tau = 10$, EASGD is found to perform better than both All-reduce SGD and Gossiping SGD. Also noted is that All-reduce SGD is found to be as fast as Gossiping SGD, by number of training epochs as well as wall clock time. All methods are found to converge to the same minimum loss at this cluster size. With 16 processes and $\tau = 10$, Gossiping SGD is found to converge faster than EASGD, which in turn converges faster than All-reduce SGD. At larger cluster sizes of 32, 64, and 128 processes, EASGD is found not to perform as well as Gossiping SGD or All-reduce SGD. Gossip is found to converge slower at smaller step sizes than All-reduce SGD. \\

\citep{jin2016gossip} notably find that distinct behaviors are exhibited based on cluster-size. With a smaller number of processes (16-32), the asynchronous methods EASGD and Gossiping SGD converge faster than All-reduce SGD, but at a large number of processes (128), All-reduce SGD is found to achieve better performance measured by validation accuracy after convergence. This result, combined with the cluster-size independent scaling of ring-reduce, provides a compelling reason to use All-reduce SGD at scale. Note again, however, that this behavior of EASGD and Gossiping SGD may not be reproducible, owing to asynchrony, and may thus perform differently under alternate test conditions. \\

\citet{blot2016gossip} compare GoSGD against EASGD on the CIFAR-10 task. The results presented are only for training loss, but interestingly show faster convergence than EASGD at both $p = 1.0$ and at $p = 0.02$, the probability that a worker engages in a gossip exchange. Due to the absence in reporting of more exhaustive experimentation, it is not possible to infer much about GoSGD's performance and viability. \\

All of these results and the discussions presented by the respective authors seem to indicate that performance of various techniques and associated hyper-parameters may vary based on problem and domain. It may also warrant a more rigorous search for optimal hyper parameters. The number of variables affecting performance was one of the primary motivating factors to restrict experimentation in this thesis to synchronous settings, so as to at least understand the behavior of various approaches in the absence of extraneous factors.

%% file: chapters/3_elastic_gossip.tex
\chapter{Elastic Gossip}
\thispagestyle{plain}
\label{ch:elastic-gossip}

Elastic Gossip, introduced in this thesis, is an extension of EASGD \citep{zhang2015easgd} which does not make use of a central process. Instead, an alternate formulation is derived using a variant of the \textit{global consensus problem} that only uses local variables. The notion of consensus in Elastic Gossip is realized through pairwise (p2p) communication, as is common with gossip-like protocols.

\section{Formulation}
The global consensus problem as discussed by \citet{zhang2015easgd} and \citet{boyd2011admm} can be used to split a single global objective function into a sum of multiple parts, where the optimization of each part may be managed by a single worker process in a distributed system. This is shown in Equation \ref{eq:global-consensus-1}.

\begin{equation}
	\label{eq:global-consensus-1}
	\min_{\theta} f(\theta) = \sum_{i \in W} f^i(\theta) \\
\end{equation}

\noindent where $f$ is the learning objective, $\theta$ are the model parameters, $W$ is the set of worker processes, and $f$ can be linearly decomposed into $f^i$. \\

If each worker were to maintain it's own set of parameters, subject to the constraint that all of them were equal, then the equivalent formulation is shown in Equation \ref{eq:global-consensus-2} \citep{boyd2011admm}. \\

\begin{equation}
	\label{eq:global-consensus-2}
	\begin{split} \min_{\theta} f(\theta) = & \sum_{i \in W} f^i(\theta^i) \\
															 & \text{subject to} \quad \theta^i - \tilde{\theta} = 0
	\end{split}
\end{equation}

\noindent where $\theta^i$ are parameters maintained at each worker $i \in W$ and $\tilde{\theta}$ is a global variable whose sole purpose is to enforce the constraint. From this, \citet{zhang2015easgd} derive the objective for EASGD, presented earlier, in Equation \ref{eq:easgd-objective}. \\

To avoid the use of a center variable, Equation \ref{eq:global-consensus-2} can equivalently be rewritten as shown in Equation \ref{eq:global-consensus-3}, and the objective derived by \citet{zhang2015easgd} (shown in Equation \ref{eq:easgd-objective}) can correspondingly be rewritten as shown in Equation \ref{eq:global-consensus-reformulation}.

\begin{equation}
	\label{eq:global-consensus-3}
	\begin{split} \min_{\theta} f(\theta) = & \sum_{i \in W} f^i(\theta^i) \\
															 & \text{subject to} \quad \sum_{k \in W} \left|\theta^i - \theta^k\right| = 0
	\end{split}
\end{equation}

\begin{equation}
	\label{eq:global-consensus-reformulation}
	\min_{\theta_i \forall i \in W} \sum_{i \in W} \left[ \: \E{f(\theta^i; \mathcal{X}^i)} + \frac{\rho}{2} \sum_{k \in W} \norm{\theta^i - \theta^k}_2^2 \: \right]
\end{equation}

\noindent where, similar to Equation \ref{eq:easgd-objective}, $W$ is the set of worker processes, $\theta^i$ are the model parameters of the workers' replicas, referred to as \textit{local variables}, and $\mathcal{X}^i$ are the partitions of training data. The first term under the summation is the training loss at each worker process, and the second term is a quadratic penalty on deviation of $\theta^i$ and from all other local variables. \\

The update rule derived subsequently for each $i \in W$ is shown in Equation \ref{eq:global-consensus-update}

\begin{equation}
	\label{eq:global-consensus-update}
	\theta^i_{t+1} = \theta^i_t - \eta \: \nabla f \! \left( \theta^i_t \right) - \alpha \sum_{k \in W} \left( \theta^i_t - \theta^k_t \right)
\end{equation}

\noindent where, consistent with Equation \ref{eq:easgd-update1}, $\eta$ is the learning rate, and $\alpha=2\eta\rho$. Note that the communication-related component in this update rule is symmetric if $\alpha$ is constant across all $i \in W$. \\

If communication is required to be restricted to pairwise interaction between workers (similar to Gossip), the summation over $k \in W$ in the communication-related component in Equation \ref{eq:global-consensus-update} may be estimated by choosing a peer $k^\prime$ uniformly, such that $\E{\theta^{k^\prime}} = \frac{1}{\left|W\right|}\sum_{k \in W}\theta^k$. The estimate is then given by Equation \ref{eq:elastic-gossip-estimate}. \\

\begin{equation}
	\label{eq:elastic-gossip-estimate}
	\sum_{k \in W} \left( \theta^i - \theta^k \right) \approx \left|W\right| \left( \theta^i - \theta^{k^\prime} \right)
\end{equation}

This is consistent with the derivation of the	 formulation used by \citep{jin2016gossip}. If we simply use this estimate in Equation \ref{eq:global-consensus-update}, we get the update rule shown in Equation \ref{eq:elastic-gossip-update-1} for $\theta^i$. By itself, this update alters $\E{\theta^{k^\prime}}$, and so correcting for this involves symmetrically adding the communication related component to $\theta^{k^\prime}$ as shown in Equation \ref{eq:elastic-gossip-update-2}. \\

\begin{equation}
	\label{eq:elastic-gossip-update-1}
	\begin{split}
	\theta^i_{t+1} & = \theta^i_t - \eta \: \nabla f \! \left( \theta^i_t \right) - \alpha \left|W\right| \left( \theta^i_t - \theta^{k^\prime}_t \right) \\
						 & = \theta^i_t - \eta \: \nabla f \! \left( \theta^i_t \right) - \alpha^\prime \left( \theta^i_t - \theta^{k^\prime}_t \right)
	\end{split}
\end{equation}

\begin{equation}
\label{eq:elastic-gossip-update-2}
\begin{split}
\theta^{k^\prime}_{t+1} = \theta^{k^\prime}_t + \alpha^\prime \left( \theta^i_t - \theta^{k^\prime}_t \right)
\end{split}
\end{equation}

\noindent where the modified moving rate $\alpha^\prime = \alpha\left|W\right|$ remains a hyper-parameter. In the rest of this thesis, the term $\alpha$ used in the context of Elastic Gossip is intended to refer to $\alpha^\prime$, and the term $k$ to $k^\prime$. \\

Equations \ref{eq:elastic-gossip-update-1} and \ref{eq:elastic-gossip-update-2} constitute the update rules for \textit{Elastic Gossip}, as introduced in this thesis, and are presented in Algorithm \ref{alg:elastic-gossip}. Note that communication is restricted to pairwise interaction as is common with Gossip-based protocols, and we delay communication to every $\tau$ updates as is common practice \citep{zhang2015easgd, jin2016gossip, blot2016gossip, strom2015quantized}. \\

\begin{algorithm}
	\caption{Elastic Gossip}
	\label{alg:elastic-gossip}
	\begin{algorithmic}
		\State \textbf{inputs:}
			\State \quad $\mathcal{X}$ \Comment{\parbox[t]{.8\linewidth}{training data instances}}
			\State \quad $\theta$ \Comment{\parbox[t]{.8\linewidth}{initial parameters of the model}}
			\State \quad $f(\theta; \mathcal{X})$ \Comment{\parbox[t]{.8\linewidth}{loss function}}
			\State \quad $W$ \Comment{\parbox[t]{.8\linewidth}{set of worker processes}}
			\State \quad $\eta$ \Comment{\parbox[t]{.8\linewidth}{learning rate}}
			\State \quad $\alpha$ \Comment{\parbox[t]{.8\linewidth}{moving rate}}
			\State \quad $\tau$ \Comment{\parbox[t]{.8\linewidth}{communication period}}
		\State \textbf{initialize}
			\State \quad $\theta^i \leftarrow \theta$, for all $i \in W$
			\State \quad $t^i \leftarrow 0$, for all $i \in W$
	\end{algorithmic}
	\hrulefill
	\begin{algorithmic}[1]
		\Repeat \: for all $i \in W$ in parallel
			\State $g^i \leftarrow \nabla f(\theta^i; x \small{\sim} \mathcal{X}^i)$
				\Comment{\parbox[t]{.53\linewidth}{compute gradients using the current state of parameters}}
			\If{$\tau$ divides $t^i$}
				\State Wait until $t^i = t^j$, for all $j \in W, i \ne j$	
					\Comment{\parbox[t]{.33\linewidth}{synchronize}}\label{alg-line:elastic-gossip-synchronize}
				\State $k^\prime \sim W \setminus \{i\}$
					\Comment{\parbox[t]{.53\linewidth}{select a peer at random}}
				\State $\theta^i \leftarrow \theta^i - \alpha \sum_{k \in \mathcal{K}} (\theta^i - \theta^k)$
					\Comment{\parbox[t]{.53\linewidth}{update parameters, where $\mathcal{K}$ is a set of processes that includes $k^\prime$ and those that selected $i$ as their peer}}
			\EndIf
			\State $\theta^i \leftarrow \theta^i - \eta g^i$
				\Comment{\parbox[t]{.53\linewidth}{update parameters}}
			\State $t^i \leftarrow t^i + 1$
				\Comment{\parbox[t]{.53\linewidth}{update clock}}
		\Until forever
	\end{algorithmic}
\end{algorithm}

\section{Relationship with EASGD and Gossiping SGD}
Note that besides the communication-related component, Elastic Gossip is identical to EASGD (Algorithm \ref{alg:easgd-sync}) \citep{zhang2015easgd} and Gossiping SGD (Algorithms \ref{alg:pull-gossiping-sgd}, \ref{alg:push-gossiping-sgd}) \citep{jin2016gossip} as presented in Chapter \ref{ch:related-work}. \\

Further, Equation \ref{eq:global-consensus-update} may be thought of as a generalization of all three approaches. A constant $\alpha$ across all $k \in W$ enforces ``elastic symmetry''. If communication is restricted to pairwise interaction between workers, then Elastic Gossip would be derived. If instead of pairwise communication, one worker process is designated as the sole point of contact for all other workers, and is not assigned a training data partition, the resulting formulation would be equivalent to EASGD. With pairwise communication, if the restriction of maintaining a constant $\alpha$ across workers and updates is relaxed, the update rules for variants of Gossiping SGD and GoSGD may be derived. \\

\section{The moving rate}
\label{sec: moving-rate}
As with EASGD, the moving rate, $\alpha$ determines the degree to which local variables can deviate from the consensus, and from each other. At one extreme, $\alpha=0$ results in workers effectively not communicating with each other throughout the training process. At the other extreme, $\alpha=1$ results in each worker replacing its local variable with that of its peer every time they communicate. $\alpha=0.5$ results in each worker setting its local variable to the average of its local variable and its peer's local variable. If we were to ignore the gradient-related component, then Equation \ref{eq:moving-rate} illustrates this behavior. \\

\begin{equation}
	\label{eq:moving-rate}
	\theta^i = \begin{cases} 
			\theta^i & \alpha = 0.0 \\
			\left(\theta^i + \theta^k\right)/2 & \alpha = 0.5 \\
			\theta^k & \alpha = 1.0 
	\end{cases}
\end{equation}

\noindent where $\theta^i$ and $\theta^k$ are a worker's local variable and its peer's local variable respectively. \\

Intuitively, Elastic Gossip may be thought of as a set of worker processes exploring a parameter space dotted with several local attractors (local optima), such that the workers are held together by an elastic (symmetric) force. $\alpha$ would then be analogous to the Elastic Modulus, determining the degree of ``strain'' (distance between workers) permissible for a given value of ``stress'' induced by forces of attraction originating from local attractors. \\

\section{Elastic Gossip Architecture}
As with EASGD \citep{zhang2015easgd}, Gossiping SGD \citep{jin2016gossip} and GoSGD \citep{blot2016gossip}, Elastic Gossip uses a data-parallel architecture where training data is partitioned across multiple worker processes, with each of them also receiving a replica of the model to be trained. As the formulation does not make use of a center variable, there is no need for a central coordinating process and thus the architecture may be deemed \textit{decentralized}. \\

While the algorithm presented here is intended for use in a synchronous setting, it can trivially be extended to the asynchronous setting by simply dropping the synchronization step in Line \ref{alg-line:elastic-gossip-synchronize} of Algorithm \ref{alg:elastic-gossip}, so long as all communicating pairs are ready to communicate. This is consistent with how synchronous EASGD is made asynchronous in formulation described by \citet{zhang2015easgd}. Elastic Gossip does not prescribe any specific data distribution strategies, as with EASGD \citep{zhang2015easgd}.

%% file: chapters/4_experiments.tex
\chapter{Experiments}
\thispagestyle{plain}
\label{ch:experiments}

In this chapter, we compare Elastic Gossip with Gossiping SGD, All-reduce, and a non-distributed (single-worker) setting. The comparisons with Gossiping SGD and All-reduce use various configurations of cluster size and communication probability. Only the pull-variant of Gossiping SGD is considered in these experiments since \citet{jin2016gossip} report that it performs better than the push variant. The effect of moving-rate on Elastic Gossip is also studied. \\

These experiments are based on the permutation invariant version of the MNIST digit recognition task \citep{mnist} and the CIFAR-10 image classification task \citep{cifar}. \\

The Deep Learning framework, PyTorch\footnote{https://pytorch.org/}, was used for all of the implementations as it has a very user-friendly API for routines used in distributed training at multiple levels of abstraction, it can seamlessly integrate with other supporting frameworks in Python, and has strong support and a thriving community. \\

\section{MNIST}
\label{sec:mnist}

The MNIST dataset \citep{mnist} is composed of 60,000 training instances and 10,000 test instances. Each instance is a labeled black-and-white (single-channel) image of a handwritten digit between 0 and 9, and is 28x28 pixels in dimensions. The learning task is to classify each given image according to the labeled digit. \\

The permutation invariant version of this task requires that any trained model be invariant to permutations in the input, implying that any structural information present in the images may not be utilized by the model. This eliminates the option of using convolutional neural networks. \\

The training procedure and neural network architecture used here closely resembles one of those studied by \citet{srivastava2014dropout}. It is a multi-layer perceptron utilizing the following: 

\begin{itemize}
	\item three dense (fully-connected) layers with 1024 units each
	\item Kaiming-initialization for all parameters \citep{he2015init,sutskever2013momentum}
	\item dropout of $p=0.2$ at the inputs and $p=0.5$ at each hidden layer \citep{srivastava2014dropout}
	\item ReLU activations at each hidden layer \citep{nair2010relu}
	\item ten-way Softmax classifier at the output
	\item mini-batch gradient descent using Nesterov's Accelerated Gradient method \citep{sutskever2013momentum}, modified as required for Elastic Gossip and Gossiping SGD
		\footnote{The experiments discussed in this chapter utilize variants of the algorithms described in Chapters \ref{ch:related-work} and \ref{ch:elastic-gossip} that incorporate Nesterov's Accelerated Gradient method, and substitute communication probability for communication period. These modifications are discussed in section \ref{sec:alg-mods} of the Appendix}
	\item learning rate of 0.001
	\item momentum of 0.99
	\item effective batch size of 128 instances \footnote{The effective batch size is the total number of training instances used in a single mini-batch update across all workers. For example, for an effective batch size of 128 instances with two workers, the mini-batch size at each worker would be 64.}
	\item trained to 100 epochs (or equivalently 40,000 weight updates) \footnote{Each epoch constitutes 400 weight updates given an effective batch size of 128 and a training set size of 51200.}
\end{itemize}

\noindent Since this is the permutation invariant version of the task, the input to the neural network is a one dimensional array with 784 values, each corresponding to a single pixel. A validation set of 8800 instances was sampled at random (without replacement) from the training set which was solely used to monitor training progress and not otherwise used for training the models. All images were preprocessed by subtracting the mean pixel activation, and dividing by the standard deviation as measured across all training instances, so as to have zero-mean and unit-variance \\

The pre-processing and model architecture described here is used across all experiments studied in this section. \\

\subsection{Single worker baseline}

As a baseline, this model was trained with a single worker in a non-distributed setting. With training replicated across four distinct random initializations, the final accuracy obtained on the test-set was between 98.51\% and 98.61\%\footnote{\citet{srivastava2014dropout} report the best performing neural network with near-identical architecture as achieving a test-set accuracy of 98.75\%. For a discussion on potential reasons for this discrepancy, please refer to section \ref{subsec:mnist-discrepancy} in the Appendix.}. Figure \ref{fig: mnist-single} shows the evolution of training loss, validation loss and validation accuracy. \\

\begin{figure}[htbp]
	\centering
	\begin{subfigure}[t]{0.48\textwidth}
		\centering
		\includegraphics[width=\linewidth]{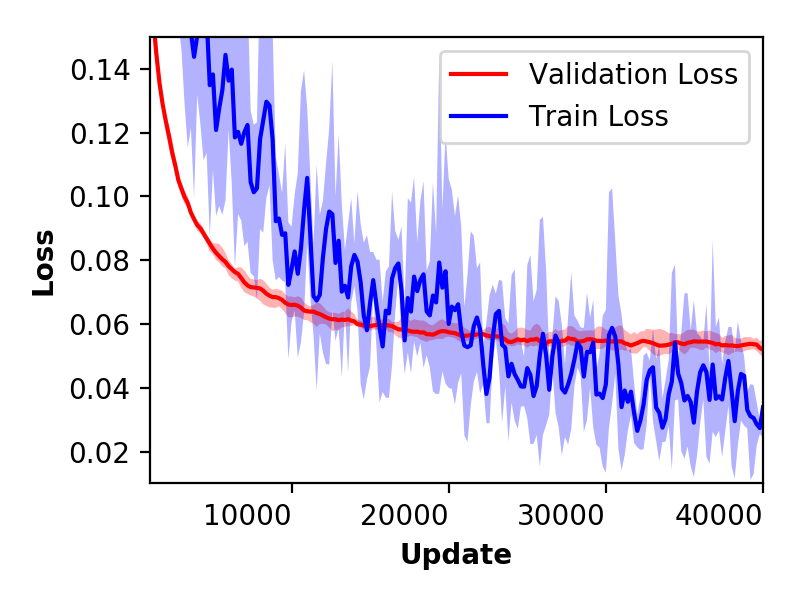}
		\caption{Training and Validation losses}
	\end{subfigure}
	\begin{subfigure}[t]{0.48\textwidth}
		\centering
		\includegraphics[width=\linewidth]{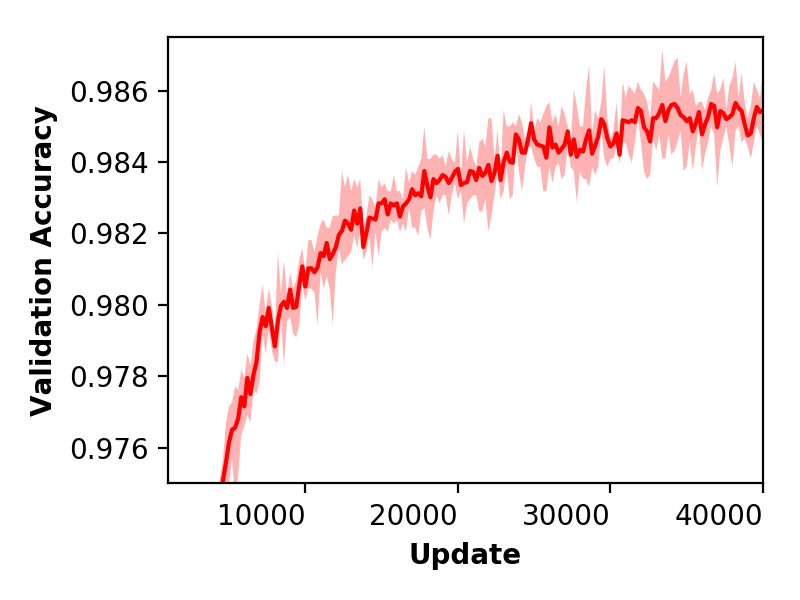}
		\caption{Validation accuracy.}
	\end{subfigure}
	\caption{Performance of models trained with SGD on a single worker (non-distributed setting), where the solid line and corresponding shaded regions indicate the mean and range, respectively, across four random initializations.}
	\label{fig: mnist-single}
\end{figure}

\subsection{Comparing All-reduce, Gossiping SGD and Elastic Gossip}

The model described was then trained using each of \textit{All-reduce}, \textit{Gossiping SGD}, and \textit{Elastic Gossip}, experimenting with 4-worker and 8-worker clusters. Table \ref{table: mnist-main-summary} summarizes the test accuracies obtained during these experiments. The \textit{Rank-0 Accuracy} and \textit{Aggregate Accuracy} are reported for each experiment. The former refers to the performance of the model trained by the Rank-0\textsuperscript{th} worker\footnote{The workers in a cluster are assigned a 0-based index known as ranks to serve as identifiers. For example, a cluster of 4 workers would be assigned ranks 0 through 3.} as measured by its accuracy on the test-set, while the latter refers to the performance of the model resulting from averaging parameters across all workers. \\

\begin{table*}[htbp]
	\centering
	\begin{tabular}{lccccc}
		\toprule[1pt]
		Method &  $\left|W\right|$ &         $p$ &          Label & \tworow[t]{Rank-0 \\ Accuracy}  & \tworow[t]{Aggregate \\ Accuracy}\\
		\midrule
		All Reduce &  4 &         - &        AR-4 &          \textbf{0.9861} &                  - \\
		No Communication &  4 &         - &        NC-4 &          0.9723 &                  - \\
		\rowcolor{table-gray}
		Elastic Gossip &  4 &  0.125000 &  EG-4-0.125 &          0.9862 &            \textbf{0.9861} \\
		\rowcolor{table-gray}
		Gossiping SGD &  4 &  0.125000 &  GS-4-0.125 &          0.9855 &             0.9850 \\
		Elastic Gossip &  4 &  0.031250 &  EG-4-0.031 &          0.9861 &             \textbf{0.9862} \\
		Gossiping SGD &  4 &  0.031250 &  GS-4-0.031 &          0.9849 &             0.9850 \\
		\rowcolor{table-gray}
		Elastic Gossip &  4 &  0.007812 &  EG-4-0.008 &          0.9838 &             \textbf{0.9853} \\
		\rowcolor{table-gray}
		Gossiping SGD &  4 &  0.007812 &  GS-4-0.008 &          0.9830 &             0.9847 \\
		Elastic Gossip &  4 &  0.001953 &  EG-4-0.002 &          0.9847 &             \textbf{0.9844} \\
		Gossiping SGD &  4 &  0.001953 &  GS-4-0.002 &          0.9823 &             0.9829 \\
		\rowcolor{table-gray}
		Elastic Gossip &  8 &  0.031250 &  EG-8-0.031 &          0.9845 &             \textbf{0.9854} \\
		\rowcolor{table-gray}
		Gossiping SGD &  8 &  0.031250 &  GS-8-0.031 &          0.9838 &             0.9842 \\
		Elastic Gossip &  8 &  0.007812 &  EG-8-0.008 &           0.9850 &             \textbf{0.9852} \\
		Gossiping SGD &  8 &  0.007812 &  GS-8-0.008 &           0.9820 &             0.9824 \\
		\rowcolor{table-gray}
		Elastic Gossip &  8 &  0.001953 &  EG-8-0.002 &          0.9772 &             \textbf{0.9812} \\
		\rowcolor{table-gray}
		Gossiping SGD &  8 &  0.001953 &  GS-8-0.002 &          0.9767 &             0.9778 \\
		\bottomrule[1pt]
	\end{tabular}
	\caption{Comparison of various configurations in terms of training method, number of workers, communication probability in case of Gossiping SGD and Elastic Gossip; $\left|W\right|$ is the number of worker processes, $p$ is the probability that a worker engages in gossip. In each experiment involving Elastic Gossip, $\alpha$ was set at 0.5. The \textit{Label}s associated with each experiment serve as identifiers that are referenced from related figures. Each of these experiments are initialized with the same random seed.}
	\label{table: mnist-main-summary}
\end{table*}

The method \textit{No Communication} involves an experiment where there was no communication between the workers throughout the training process, which in effect resulted in each worker learning a model given only the data partition it was assigned. The performance of these models may be thought of as a lower bound. \\

\begin{figure}[htbp]
	\centering
	\includegraphics[width=\linewidth]{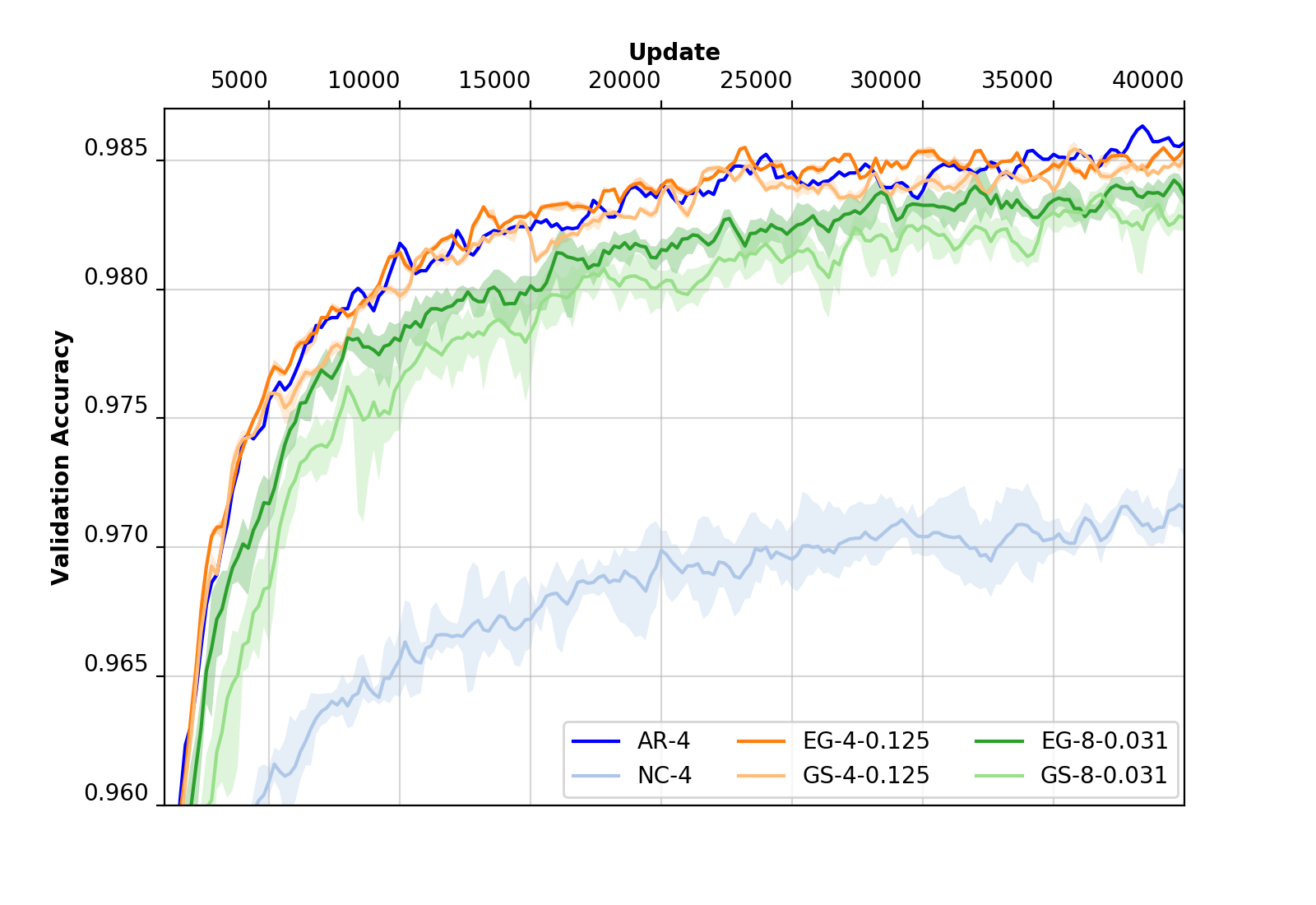}
	\caption{Evolution of validation accuracy for some of the experiments from Table \ref{table: mnist-main-summary}. Each pair of solid line and associated shaded region represents the average and range, respectively, of validation accuracies across all workers at the given epoch. The labels in the legend correspond to labels in Table \ref{table: mnist-main-summary}.}
	\label{fig: mnist-methods-best}
\end{figure}

\begin{figure}[htbp]
	\centering
	\includegraphics[width=\linewidth]{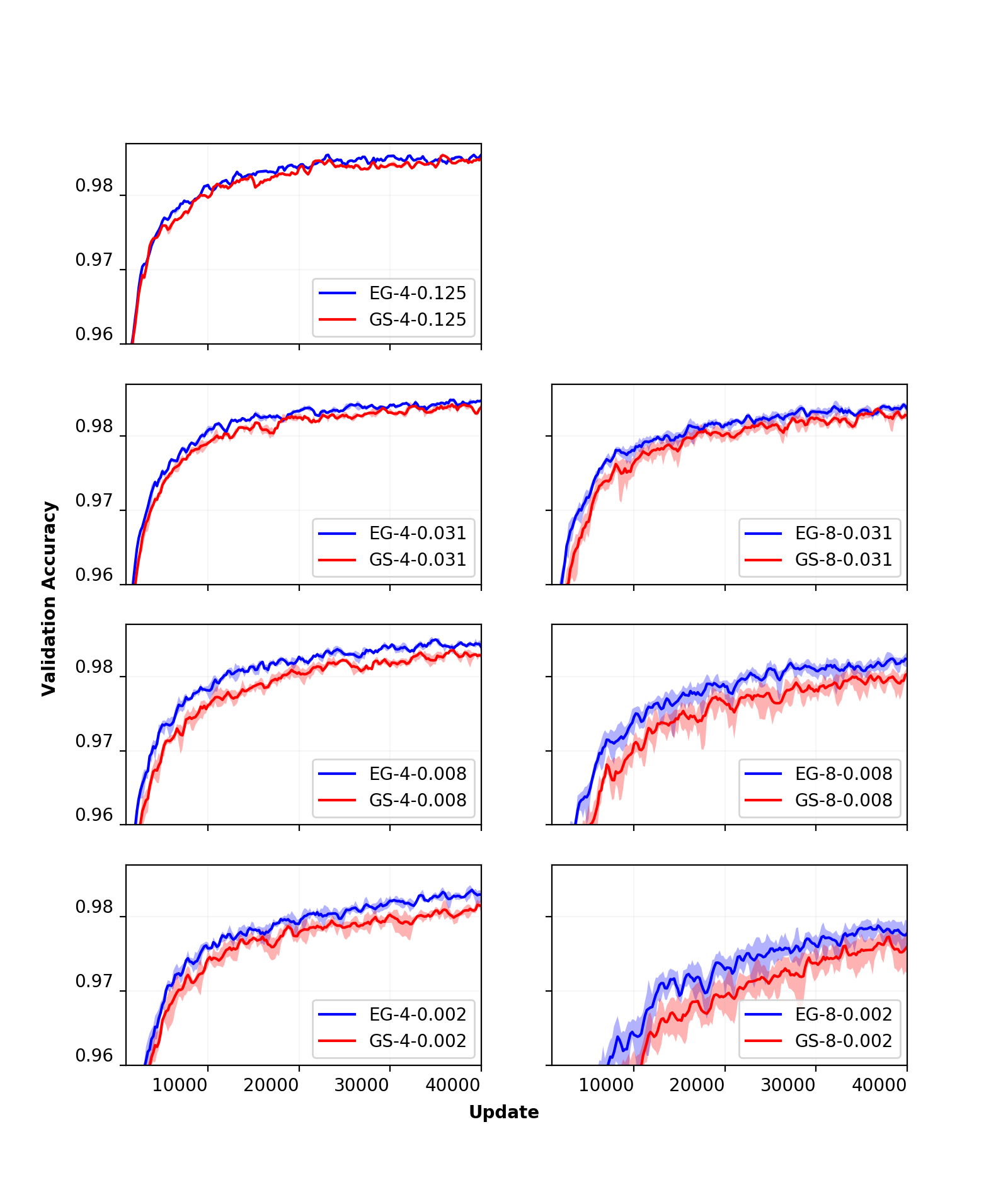}
	\caption{Evolution of validation accuracy comparing Elastic Gossip with Gossiping SGD over a range of values for communication probability and number of workers. Each pair of solid line and associated shaded region represents the average and range, respectively, of validation accuracies across all workers at the given epoch. The labels in the legend correspond to labels in Table \ref{table: mnist-main-summary}. In general, the blue curves correspond to Elastic Gossip and the red curves to their Gossiping SGD counterparts.}
	\label{fig: mnist-methods-eg-v-gs}
\end{figure}

Figure \ref{fig: mnist-methods-best} shows the evolution of validation losses for some of these experiments that have comparable results. Figure \ref{fig: mnist-methods-eg-v-gs} shows a comparison between Elastic Gossip and Gossiping SGD for a range of values of communication probability and number of workers. \\

From these results, Elastic Gossip appears to consistently outperform Gossiping SGD. Since the only distinction between the Gossiping SGD case and Elastic Gossip with $\alpha=0.5$ is that communication in the latter approach is bidirectional, the improved performance potentially comes at an added communication cost. However, it is to be noted that Elastic Gossip with a lower communication probability may perform better than Gossiping SGD, considering, for example, that the experiment labeled EG-4-0.031 results in significantly better performance than GS-4-0.125, despite the former's communication probability being less than the latter's by a factor of 4. \\

It is also interesting to note that at higher values of $p$, the performance of All-reduce and Elastic Gossip are very similar, both in terms of test-set accuracies as well as evolution of validation accuracies, while Gossiping SGD is also a close contender. This indicates that synchronous distributed training at the scale studied may be conducted at much lower communication overhead than is standard practice with All-reduce. \\

\subsection{The effect of moving rate}
As discussed in Section \ref{sec: moving-rate}, one of the primary advantages that Elastic Gossip provides over Gossiping SGD is the moving rate hyper-parameter, $\alpha$. This hyper-parameter provides the ability to control the explore-exploit tradeoff, such that a lower $\alpha$ results in greater deviation of workers' parameters (local variables) from each other, thus enabling a higher degree of ``exploration'' in parameter space. \\

\begin{figure}[htbp]
	\centering
	\includegraphics[width=0.75\linewidth]{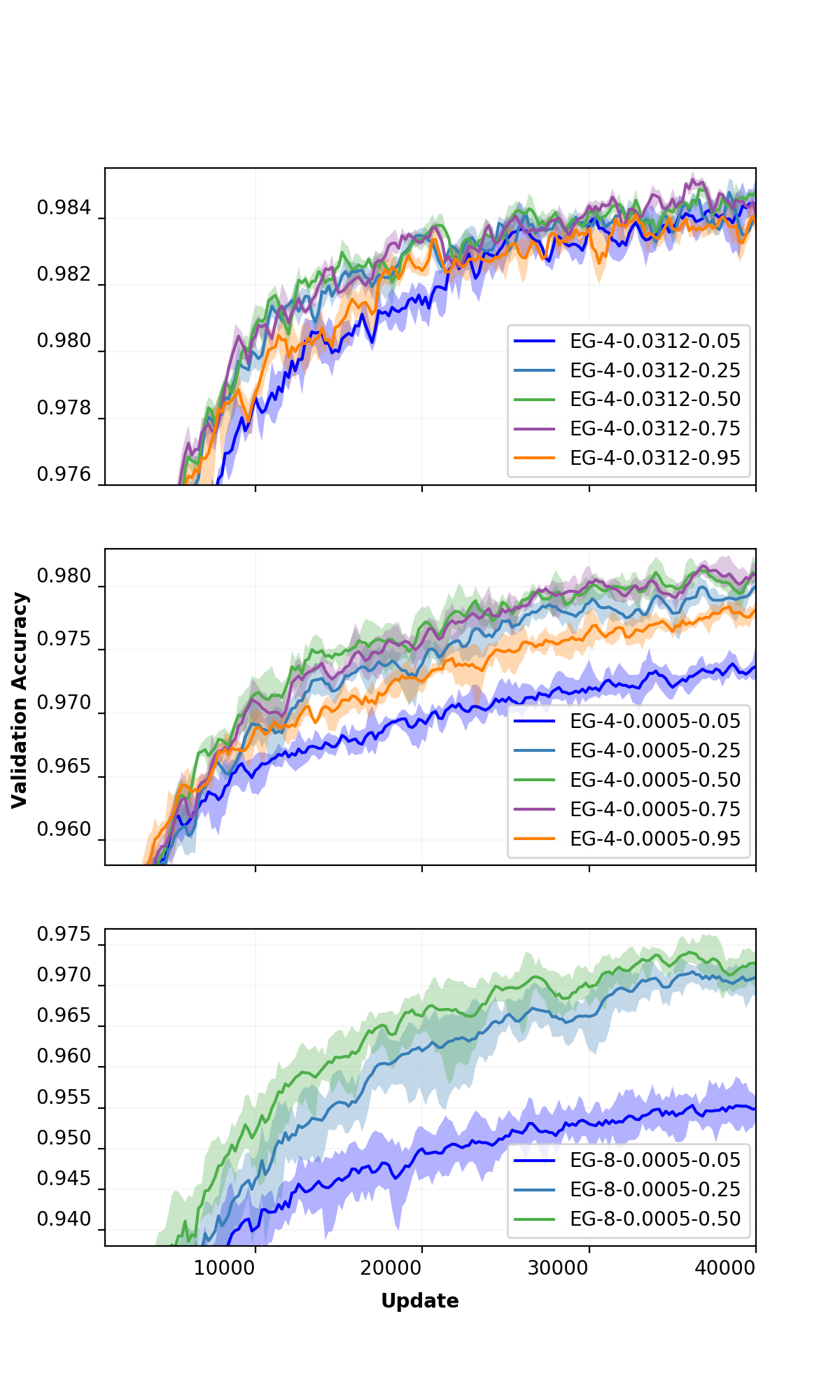}
	\caption{The effect of moving rate, $\alpha$, on training using Elastic Gossip. Please note that the y-axes have differing scales.}
	\label{fig: mnist-alpha}
\end{figure}

A few experiments aimed at studying the effects of $\alpha$ at various values of $p$ and $\left|W\right|$ are summarized in Table \ref{tab: mnist-alpha}. Figure \ref{fig: mnist-alpha} shows the evolution of validation-set accuracies in these experiments. \\

\begin{table*}[htbp]
	\centering
\begin{tabular}{cccccc}
	\toprule
	$\left|W\right|$ &         $p$ &  $\alpha$&        Label & \tworow[t]{Rank-0 \\ Accuracy}  & \tworow[t]{Aggregate \\ Accuracy}\\
	\midrule
	 4 &  0.031250 &   0.05 &  EG-4-0.0312-0.05 &          0.9833 &              0.9850 \\
	4 &  0.031250 &   0.25 &  EG-4-0.0312-0.25 &           0.9860 &             \textbf{0.9865} \\
	4 &  0.031250 &   0.50 &  EG-4-0.0312-0.50 &          0.9861 &             0.9862 \\
	4 &  0.031250 &   0.75 &  EG-4-0.0312-0.75 &          0.9846 &              0.9850 \\
	4 &  0.031250 &   0.95 &  EG-4-0.0312-0.95 &          0.9846 &             0.9857 \\
	\rowcolor{table-gray}
	4 &  0.000488 &   0.05 &  EG-4-0.0005-0.05 &          0.9752 &             0.9647 \\
	\rowcolor{table-gray}
	4 &  0.000488 &   0.25 &  EG-4-0.0005-0.25 &          0.9816 &             0.9826 \\
	\rowcolor{table-gray}
	4 &  0.000488 &   0.50 &  EG-4-0.0005-0.50 &          0.9814 &             \textbf{0.9834} \\
	\rowcolor{table-gray}
	4 &  0.000488 &   0.75 &  EG-4-0.0005-0.75 &          0.9813 &             0.9825 \\
	\rowcolor{table-gray}
	4 &  0.000488 &   0.95 &  EG-4-0.0005-0.95 &          0.9801 &             0.9765 \\
	8 &  0.000488 &   0.05 &  EG-8-0.0005-0.05 &          0.9532 &             \textbf{0.4309} \\
	8 &  0.000488 &   0.25 &  EG-8-0.0005-0.25 &          0.9719 &             0.9708 \\
	8 &  0.000488 &   0.50 &  EG-8-0.0005-0.50 &          0.9722 &             \textbf{0.9747} \\
	\bottomrule
\end{tabular}
	\caption{The effect of moving rate, $\alpha$, on training using Elastic Gossip. The labels listed here reference plots in Figure \ref{fig: mnist-alpha}}
	\label{tab: mnist-alpha}
\end{table*}

While choosing $\alpha=0.5$ seems like a safe choice for these experiments, and performance seems to generally get worse monotonically as $\alpha$ is either increased or decreased, the results do not appear to be conclusive and warrant a more rigorous search for optimal values for $\alpha$. For example, in the ``EG-4-0.0312-'' set of experiments, it appears that $\alpha=0.95$ works significantly better than $\alpha=0.75$. \\

Further, from the evolution of validation-set accuracies, it appears that high values of $\alpha$ tend to initially work well sometimes but results in degraded performance at later stages in training. So it may be that a schedule for changing $\alpha$ based on training stage may be more optimal than using a constant $\alpha$ throughout training, as is common practice with other hyper-parameters such as learning rate and momentum. \\

\section{CIFAR-10}

The CIFAR-10 dataset is composed of 50,000 training instances and 10,000 test instances. Each instance is a color image with 3 channels, each channel being 32x32 pixels in dimensions. Every instance is labeled as one of \textit{airplane}, \textit{automobile}, \textit{bird}, \textit{cat}, \textit{deer}, \textit{dog}, \textit{frog}, \textit{horse}, \textit{ship}, or \textit{truck}. The classification task is to label each given instance with the specified label. \\

This set of experiments train a model using the Resnet-18 architecture in its pre-activation variant \citep{he2016resnet, he2016preactivation}. The Resnet-18, despite being one of the shallower modern convolutional neural network architectures, consists of common components such as residual units \citep{he2016resnet} and batch normalization \citep{ioffe2015bn}, and is relatively faster to train than networks of similar width that are 100 or 1000 layers deep. Hence this architecture was chosen as the intention of these experiments is to study the efficacy of Elastic Gossip on common training techniques rather than push the state-of-the-art in terms of classification performance. \\

Following were the hyper-parameters used:
\begin{itemize}
	\item mini-batch gradient descent using Nesterov's Accelerated Gradient method \citep{sutskever2013momentum}
	\item initial learning rate of 0.01 annealed after 15, 30, and 40 epochs by a factor of 0.5
	\item momentum of 0.9
	\item moving-rate of 0.5 when using Elastic Gossip
	\item effective batch size of 128 instances
	\item trained to 50 epochs (or equivalently 17,500 weight updates) \footnote{Each epoch constitutes 350 weight updates given an effective batch size of 128 and a training set size of 44800.}
\end{itemize}

All instances were pre-processed to have zero-mean and unit-variance as measured on the training set. A validation set of 5200 instances was sampled from the training set and were used solely for purposes of monitoring training. 

\begin{table}[htbp]
	\centering
	\begin{tabular}{lcccc}
		\toprule
		Method &  $\left|W\right|$ &         $p$ &  \tworow[t]{Rank-0 \\ Accuracy}  & \tworow[t]{Aggregate \\ Accuracy}\\
		\midrule
		All Reduce &  4 &         - &           0.9193 &              \textbf{0.9193} \\
		\rowcolor{table-gray}
		Elastic Gossip &  4 &  0.125000 &           0.9166 &              \textbf{0.9146} \\
		\rowcolor{table-gray}
		Gossiping SGD &  4 &  0.125000 &           0.9131 &              0.9135 \\
		Elastic Gossip &  4 &  0.031250 &           0.9122 &              \textbf{0.9139} \\
		Gossiping SGD &  4 &  0.031250 &           0.9048 &              0.9065 \\
		\rowcolor{table-gray}
		Elastic Gossip &  4 &  0.007812 &           0.9006 &              0.9044 \\
		\rowcolor{table-gray}
		Gossiping SGD &  4 &  0.007812 &           0.9015 &              \textbf{0.9050} \\
		Elastic Gossip &  4 &  0.001953 &           0.8952 &              \textbf{0.8983} \\
		Gossiping SGD &  4 &  0.001953 &           0.8825 &              0.8845 \\
		\bottomrule
	\end{tabular}
	\caption{Summary of experiments conducted on the CIFAR-10 dataset. As with the experiments summarized in \ref{table: mnist-main-summary}, all experiments using Elastic Gossip have a moving rate $\alpha=0.5$}
	\label{tab: cifar-summary}
\end{table}

Table \ref{tab: cifar-summary} summarizes a set of experiments conducted using the architecture described above, varying the communication probability. From these results, it appears that Elastic Gossip again generally performs better than Gossiping SGD at this scale.

%% file: chapters/5_conclusion.tex
\chapter{Conclusion}
\thispagestyle{plain}
\label{ch:conclusion}

This thesis introduced Elastic Gossip, a decentralized extension of EASGD, for training neural networks in a distributed setting. Elastic Gossip was empirically shown to be at least comparable to Gossiping SGD, which is the most similar technique for decentralized training, both with multi-layer-perceptrons as well as modern convolutional neural networks. \\

The approach of evaluating training in a synchronous setting is a departure from standard practice in that, techniques are conventionally evaluated in asynchronous settings in an attempt to scale, but are effectively irreproducible due to extraneous factors affecting asynchronous computing. It is to be stressed that while the algorithms proposed here were defined as being synchronous, it was also shown that Elastic Gossip can be trivially extended to the asynchronous setting. \\

Elastic Gossip, besides seemingly performing better than Gossiping SGD in terms of classification accuracy, extends the notion of \textit{moving rate}, originally introduced for EASGD by \citet{zhang2015easgd}, to distributed/p2p settings. The \textit{moving rate} provides the ability to control the explore-exploit trade-off. \\

While results presented in this thesis seem very encouraging, applying Elastic Gossip (as with any other technique) would require a rigorous hyper-parameter search, and its performance in comparison to similar techniques would likely depend on the application. The results here, however, do strongly suggest that Elastic Gossip might work better than Gossiping SGD, and by extension, better than EASGD or All-reduce under certain conditions \citep{jin2016gossip}. \\

The work presented in this thesis also opens up avenues for further research and investigation. The behavior of the techniques presented here remain to be studied in environments with simulated (controlled) asynchrony. The scale of experiments presented here were relatively small, with small cluster sizes. As \citet{jin2016gossip} noted in their study, different training techniques exhibit differing behaviors based on cluster size. Whether the promise of Elastic Gossip holds at other scales remains to be studied. However, given how similar Elastic Gossip and Gossiping SGD are, one might expect similar scaling behaviors. \\

The experiments here assume homogeneous hardware and environment, and fully connected network topologies with a constant communication cost between all peers, which is generally applicable when these aspects are transparent to the user as with cloud computing or cluster computing. It will be interesting to understand distributed training behaviors in non-homogeneous environments as is common with inherently distributed systems such as IOT devices and sensor networks. \\

Considering one of the motivations for distributed training is to collocate training with data collection, data partitioning is bound to be biased and skewed as is data collection. Hence these algorithms need to be studied under such conditions of partitioning as well. \\

The experiments used here were all in the supervised learning paradigm. It would be interesting to study these techniques' capabilities in the unsupervised and reinforcement learning paradigms. It would also be interesting to see if these techniques open up alternate forms of training, for example, by treating worker processes not just as computational abstractions but as agents in a game theoretic framework.

%% file: chapters/appendix.tex
\chapter{APPENDIX}
\thispagestyle{plain}

\label{ch:appendix}

\section{Modifications to Algorithms}
\label{sec:alg-mods}

The algorithms discussed in Chapters \ref{ch:related-work} and \ref{ch:elastic-gossip} illustrate the basic formulations of Elastic Gossip and Gossiping SGD. The experiments discussed in Chapter \ref{ch:experiments} utilize modified versions of these algorithms, which are discussed here.

\subsection{Incorporating Nesterov's Accelerated Gradient Method (NAG)}

As discussed in the earlier chapters, each update of EASGD, Gossiping SGD and Elastic Gossip can be decomposed into their corresponding communication-related and gradient-related components. The differences between each of these techniques lie solely in the communication-related component. Incorporation of NAG requires a modification of the gradient-related component which is consistent across each of these techniques. Algorithm \ref{alg:elastic-gossip-nag} describes this variant for Elastic Gossip. This differs from the Algorithm \ref{alg:elastic-gossip} specifically in lines \ref{alg-line:elastic-gossip-nag-compute-nag} and \ref{alg-line:elastic-gossip-nag-gradient-update}, involving computing the velocity component and using this in the gradient-related update. A corresponding modification to Gossiping SGD and EASGD may be inferred from Algorithm \ref{alg:elastic-gossip-nag}. \\

\begin{algorithm}
	\caption{Elastic Gossip incorporating NAG and Communication Probability}
	\label{alg:elastic-gossip-nag}
	\begin{algorithmic}
		\State \textbf{inputs:}
		\State \quad $\mathcal{X}$ \Comment{\parbox[t]{.8\linewidth}{training data instances}}
		\State \quad $\theta$ \Comment{\parbox[t]{.8\linewidth}{initial parameters of the model}}
		\State \quad $f(\theta; \mathcal{X})$ \Comment{\parbox[t]{.8\linewidth}{loss function}}
		\State \quad $W$ \Comment{\parbox[t]{.8\linewidth}{set of worker processes}}
		\State \quad $\eta$ \Comment{\parbox[t]{.8\linewidth}{learning rate}}
		\State \quad $\mu$ \Comment{\parbox[t]{.8\linewidth}{momentum}}
		\State \quad $\alpha$ \Comment{\parbox[t]{.8\linewidth}{moving rate}}
		\State \quad $\tau$ \Comment{\parbox[t]{.8\linewidth}{communication period}}
		\State \textbf{initialize}
		\State \quad $\theta^i \leftarrow \theta$, for all $i \in W$
		\State \quad $t^i \leftarrow 0$, for all $i \in W$
		\State \quad $v^i \leftarrow 0$, for all $i \in W$
	\end{algorithmic}
	\hrulefill
	\begin{algorithmic}[1]
		\Repeat \: for all $i \in W$ in parallel
			\State $g^i \leftarrow \nabla f(\theta^i; x \small{\sim} \mathcal{X}^i)$
				\Comment{\parbox[t]{.53\linewidth}{compute gradients using the current state of parameters}}
			\State $v^i \leftarrow \mu v^i - \eta g^i$
				\Comment{\parbox[t]{.53\linewidth}{compute velocity (based on NAG)}}\label{alg-line:elastic-gossip-nag-compute-nag}
			\If{$True \sim Bernoulli(p)$}\label{alg-line:elastic-gossip-nag-probability}
				\State Wait until $t^i = t^j$, for all $j \in W, i \ne j$	
					\Comment{\parbox[t]{.33\linewidth}{synchronize}}
				\State $k^\prime \sim W \setminus \{i\}$
					\Comment{\parbox[t]{.53\linewidth}{select a peer at random}}
				\State $\theta^i \leftarrow \theta^i - \alpha \sum_{k \in \mathcal{K}} (\theta^i - \theta^k)$
					\Comment{\parbox[t]{.53\linewidth}{update parameters, where $\mathcal{K}$ is a set of processes that includes $k^\prime$ and those that selected $i$ as their peer}}
			\EndIf
			\State $\theta^i \leftarrow \theta^i - \eta g^i  + \mu v^i$
				\Comment{\parbox[t]{.53\linewidth}{update parameters}}\label{alg-line:elastic-gossip-nag-gradient-update}
			\State $t^i \leftarrow t^i + 1$
				\Comment{\parbox[t]{.53\linewidth}{update clock}}
			\Until forever
	\end{algorithmic}
\end{algorithm}

\subsection{Communication Probability}

The algorithms described thus far utilize a communication period, $\tau$, such that communication is restricted to every $\tau$ updates instead of every single update. The experiments discussed in Chapter \ref{ch:experiments}, however, utilize a communication probability $p$, such that each process decides to communicate in a given iteration with probability $p$ of sampling $True$ from a Bernoulli Distribution, and the communication period is then $1/p$ in expectation. This is similar to the formulation proposed by \citet{blot2016gossip} for GoSGD. Line \ref{alg-line:elastic-gossip-nag-probability} of Algorithm \ref{alg:elastic-gossip-nag} shows the incorporation of communication probability. \\

\begin{table}
	\centering
\begin{tabular}{ccccc}
	\toprule
	$p$ &  $\tau_{eff}$ &    $\tau$ & \tworow[t]{Rank-0 \\ Accuracy}  & \tworow[t]{Aggregate \\ Accuracy}\\
	\midrule
	- &        - &    8 &          0.9864 &             0.9865 \\
	0.125000 &      8 &      - &          0.9855 &              0.9850 \\
	\rowcolor{table-gray}
	- &        - &   32 &          0.9857 &             0.9858 \\
	\rowcolor{table-gray}
	0.031250 &     32 &      - &          0.9849 &              0.9850 \\
	- &        - &  128 &          0.9846 &             0.9848 \\
	0.007812 &    128 &      - &           0.9830 &             0.9847 \\
	\rowcolor{table-gray}
	- &        - &  512 &          0.9833 &             0.9843 \\
	\rowcolor{table-gray}
	0.001953 &    512 &      - &          0.9823 &             0.9829 \\
	\bottomrule
\end{tabular}
\caption{Comparison between performance of models trained using various values of communication probability $p$, and communication period $\tau$, using Gossiping SGD with 4 workers on the MNIST classification task, where $\tau_{eff} = \frac{1}{p}$ is the effective communication period.}
\label{table: communication-period-probability}
\end{table}

The primary intended advantage of using communication probability is to prevent all workers from communicating simultaneously and thereby impacting constraints on the cluster network. It is however noted from experiments with Gossiping SGD that this may result in degraded performance as summarized in Table \ref{table: communication-period-probability}.

\section{Notes on the experiments}

\subsection{Explaining MNIST reproduction discrepancies}
\label{subsec:mnist-discrepancy}
As mentioned in section \ref{sec:mnist}, the performance of the baseline model as measured by accuracy on the test-set - between 98.51\% and 98.61\% across four random seeds - is lower than that reported by \citet{srivastava2014dropout} - 98.75\% - for a near identical architecture. Following are some postulated reasons that might explain the discrepancy:
\begin{enumerate}
	\item They report accuracy of the best-performing neural network only
	\item They train on the entire training-set, while we hold out 8800 instances for validation
	\item They train for 1,000,000 weight updates, but we stop training at 40,000
\end{enumerate}

\section{Push variant of Gossiping SGD}

The synchronous version of the \textit{push} variant of Gossiping SGD, originally proposed by \citet{jin2016gossip} in the asynchronous setting, is presented here in Algorithm \ref{alg:push-gossiping-sgd}.

\begin{algorithm}
	\caption{Synchronous Push-Gossiping SGD}
	\label{alg:push-gossiping-sgd}
	\begin{algorithmic}
		\State \textbf{inputs:}
		\State \quad $\mathcal{X}$ \Comment{\parbox[t]{.8\linewidth}{training data instances}}
		\State \quad $\theta$ \Comment{\parbox[t]{.8\linewidth}{initial parameters of the model}}
		\State \quad $f(\theta; \mathcal{X})$ \Comment{\parbox[t]{.8\linewidth}{loss function}}
		\State \quad $W$ \Comment{\parbox[t]{.8\linewidth}{set of worker processes}}
		\State \quad $\eta$ \Comment{\parbox[t]{.8\linewidth}{learning rate}}
		\State \quad $\tau$ \Comment{\parbox[t]{.8\linewidth}{communication period}}
		\State \textbf{initialize}
		\State \quad $\tilde{\theta} \leftarrow \theta$
		\State \quad $\theta^i \leftarrow \theta$, for all $i \in W$
		\State \quad $t^i \leftarrow 0$, for all $i \in W$
	\end{algorithmic}
	\hrulefill
	\begin{algorithmic}[1]
		\Repeat \: for all $i \in W$ in parallel
		\State $g^i \leftarrow \nabla f(\theta^i; x \small{\sim} \mathcal{X}^i)$
		\Comment{\parbox[t]{.53\linewidth}{compute gradients using the current state of parameters}}
		\If{$\tau$ divides $t^i$}
		\State Wait until $t^i = t^j$, for all $j \in W, i \ne j$
		\Comment{\parbox[t]{.33\linewidth}{synchronize}}
		\State $k^\prime \sim W \setminus \{i\}$
		\Comment{\parbox[t]{.53\linewidth}{select a peer at random}}
		\State $\theta^i \leftarrow \frac{1}{|\mathcal{K}|}\sum_{k \in \mathcal{K}}\theta^k$
		\Comment{\parbox[t]{.53\linewidth}{update parameters, where $\mathcal{K}$ is a set of processes that includes $i$ and every other process that selected $i$ as their peer}}
		\EndIf
		\State $\theta^i \leftarrow \theta^i - \eta g^i$
		\Comment{\parbox[t]{.53\linewidth}{update parameters}}
		\State $t^i \leftarrow t^i + 1$
		\Comment{\parbox[t]{.53\linewidth}{update clock}}
		\Until forever
	\end{algorithmic}
\end{algorithm}